\pdfoutput=1

\documentclass[11pt]{article}

\usepackage{naacl2021}

\usepackage{times}
\usepackage{latexsym}

\usepackage[T1]{fontenc}

\usepackage[utf8]{inputenc}

\usepackage{microtype}
\usepackage[ruled,vlined]{algorithm2e}
\usepackage{amsmath,amssymb,amsfonts}
\usepackage{textgreek}
\usepackage{graphicx}
\usepackage{multirow}
\graphicspath{{images/}}
\title{BERT based Transformers lead the way in Extraction of Health Information from Social Media}

\author{
    Sidharth R$^{1}$ \And Abhiraj Tiwari$^{1}$ \And Parthivi Choubey$^{1}$ \And Saisha Kashyap$^{2}$ \AND
     Sahil Khose$^{2}$ \And Kumud Lakara$^{1}$ \And Nishesh Singh$^{3}$ \And Ujjwal Verma$^{4}$ \AND
    \vspace*{-7mm} \\ \texttt{\{sidram2000, abhirajtiwari, parthivichoubey, saishakashyap8,}\\
    \texttt{sahilkhose18, lakara.kumud, singhnishesh4\}@gmail.com} \\
    \texttt{ujjwal.verma@manipal.edu}
}

\begin{document}

\maketitle
{\let\thefootnote\relax\footnote{\begin{flushleft}$^1$Dept. of Computer Science and Engineering\\
 $^2$Dept. of Information and Communication Technology\\
 $^3$Dept. of Mechanical and Manufacturing Engineering \\
 $^4$Dept. of Electronics and Communication Engineering\\
 Manipal Institute of Technology, Manipal Academy of Higher Education, Manipal, India\end{flushleft}}}

\begin{abstract}
This paper describes our submissions for the Social Media Mining for Health (SMM4H) 2021 shared tasks. We participated in 2 tasks: (1) Classification, extraction and normalization of adverse drug effect (ADE) mentions in English tweets (Task-1) and (2) Classification of COVID-19 tweets containing symptoms (Task-6). Our approach for the first task uses the language representation model RoBERTa with a binary classification head. For the second task, we use BERTweet, based on RoBERTa. Fine-tuning is performed on the pre-trained models for both tasks. The models are placed on top of a custom domain-specific pre-processing pipeline. Our system ranked first among all the submissions for subtask-1(a) with an F1-score of 61\%. For subtask-1(b), our system obtained an F1-score of 50\% with improvements up to +8\% F1 over the score averaged across all submissions. The BERTweet model achieved an F1 score of 94\% on SMM4H 2021 Task-6.
\end{abstract}

\section{Introduction}
Social media platforms are a feature of everyday life for a large proportion of the population with an estimated 4.2 billion people using some form of social media \cite{Hootsuite-2021}. Twitter is one of the largest social media platforms with 192 million daily active users \citep{nyt-twitter}. The 6th Social Media Mining for Health Applications Workshop focuses on the use of Natural Language Processing (NLP) for a wide number of tasks related to Health Informatics using data extracted from Twitter . \par
Our team, TensorFlu, participated in 2 tasks, (1) Task-1: Classification, extraction and normalization of adverse effect (AE) mentions in English tweets and (2) Task-6: Classification of COVID-19 tweets containing symptoms. A detailed overview of the shared tasks in the 6th edition of the workshop can be found in \cite{magge2021overview}. \par
The classification and extraction of Adverse Drug Effects (ADE) on social media can be a useful indicator to judge the efficacy of medications and drugs while ensuring that any side effects that previously remained unknown can be found. Thus social media can be a useful medium to judge gauge patient satisfaction and well being. \par 
According to the report in \citep{pew-usnews}, 15\% of American adults get their news on Twitter while 59\% of Twitter users get their news on Twitter itself. Thus during the spread of a pandemic like COVID-19, tracking reports by users as well as news mentions from local organizations can perform the function of tracking the spread of the disease in new regions and keep people informed. \par 
Similar to the last edition of the workshop, the top performing model \citep{klein-etal-2020-overview} for Task-1 with the highest score this year was RoBERTa \citep{liu2019roberta}. The biggest challenge while dealing with the dataset provided for this years competition was the huge class imbalance. The proposed approach handles this by the use of Random Sampling \cite{abd2013review} of the dataset during finetuning. Named Entity Recognition (NER) for the extraction of text spans was performed using the RoBERTa based model provided in the spaCy \cite{spacy} \verb|en_core_web_trf| pipeline. For the classification of tweets with COVID-19 symptoms, we used a model called BERTweet \cite{nguyen-etal-2020-bertweet} trained on 845 million English tweets and 23 million COVID-19 related English tweets as of the latest publically available version of the model. Fine-tuning was performed on the pretrained models for both tasks. Section \ref{Sec:task1} summarizes the methodology and results obtained for Task-1, while Section \ref{Sec:task6} summarizes the methodology and results for Task-6. 

\section{Task-1: Classification, extraction and normalization of adverse effect (AE) mentions in English tweets}
\label{Sec:task1}

\subsection{Sub-Task 1a: Classification}
The goal of this sub-task is to classify tweets that contain an adverse effect (AE) or also known as adverse drug effect (ADE) with the label ADE or NoADE.

\subsubsection{Data and Pre-processing}
The organizers of SMM4H provided us with a training set consisting of 18,256 tweets with 1,297 positive examples and 16,959 negative examples. Thus, the dataset has a huge class imbalance. The validation dataset has 913 tweets with 65 positive examples and 848 negative examples. \par
To overcome the class imbalance we performed random oversampling and undersampling \cite{abd2013review} on the provided dataset. The dataset was first oversampled using a sampling strategy of 0.1 i.e. the minority class was oversampled so that it was 0.1 times the size of majority class, then the resultant dataset was undersampled using a sampling strategy of 0.5 i.e. the majority class was undersampled so that the majority class was 2 times the size of minority class \par
Removal of twitter mentions, hashtags and URLs was performed, but it negatively affected the performance of the model. Hence, this pre-processing step was not performed in the final model.  
The tweets were then preprocessed using fairseq \cite{ott-etal-2019-fairseq} preprocessor which tokenizes the sentences using GPT-2 byte pair encoding\cite{radford2019language} and finally converts them into binary samples.

\subsubsection{System Description}
Fairseq's \cite{ott-etal-2019-fairseq} pretrained RoBERTa \cite{liu2019roberta} large model was used for the task with a binary classification head. The RoBERTa model was pretrained over 160GB of data from BookCorpus \cite{Zhu_2015_ICCV}, CC-News \cite{nagel-2016}, OpenWebText \cite{Gokaslan2019OpenWeb} and Stories.

\subsubsection{Experiments}
RoBERTa and BioBERT \cite{10.1093/bioinformatics/btz682} were trained for ADE classification and extensive hyperparameter tuning was carried out. The hyperparameters tested on the validation split included the learning rate, batch size, and sampling strategy of the dataset.
The RoBERTa model was trained for 6 epochs with a batch size of 8. The learning rate was warmed up for 217 steps with a weight decay of 0.1 and a peak learning rate of $10^{-5}$ for the polynomial learning rate scheduler. A dropout rate of 0.1 is used along with the Adam optimizer having (\textbeta$_1$, \textbeta$_2$)=(0.9, 0.98).\par

\begin{table*}[!htbp]
\centering
\begin{tabular}{ |c|c|c|c|c|c|c| } 
 \hline
 \textbf{S.No.} & \textbf{Model} & \textbf{Arch} & \textbf{Label} & \textbf{Precision} & \textbf{Recall} & \textbf{F1} \\ 
 \hline
 1. & \textbf{RoBERTa} & $BERT_{LARGE}$ & NoADE &  \textbf{0.99}  &  0.95  &  0.97  \\
    &  &    &   ADE   & 0.59  &   \textbf{0.92} & \textbf{0.72}    \\
    \hline
2.  & BioBERT  & $BERT_{BASE}$ & NoADE  & 0.97 & \textbf{0.99}  &  \textbf{0.98}  \\
    &  &    &   ADE   &  \textbf{0.78}   &   0.60   & 0.68  \\
    \hline
\end{tabular}
\caption{Comparing different models used for task 1a on the \textbf{Validation Set}. \textbf{RoBERTa} is chosen owing to its higher F1- score while predicting the ADE label correctly.}
\label{tab:Task1aComparison}
\end{table*}

\begin{table}[!htbp]
\centering
\resizebox{\columnwidth}{!}{\begin{tabular}{ |c|c|c|c|c|c|c| } 
 \hline
  & \textbf{Precision} & \textbf{Recall} & \textbf{F1} \\ 
 \hline
 \textbf{RoBERTa} & \textbf{0.515} & \textbf{0.752} & \textbf{0.61} \\
    \hline
 Median & 0.505 & 0.409 & 0.44 \\
    \hline
\end{tabular} }
\caption{Comparing our best-performing model to the median for task 1a.}
\label{tab:Task1aFinal}
\end{table}

\subsubsection{Results}
\label{sec:Equations}
Precision is defined as the ratio between true positives and the sum of true positives and false positives.
\begin{equation}
    \label{eqn:Precision}
    Precision=\frac{TP}{TP+FP} 
\end{equation}

Recall is defined as the ratio between true positives and the sum of true positives and false negatives.
\begin{equation}
    \label{eqn:Recall}
    Recall=\frac{TP}{TP+FN} 
\end{equation}

Our primary objective is to create a model that prevents incorrect classification of ADE tweets. A model with higher recall than precision is more desirable for us as the former tends to reduce the total number of false negatives.
F1 Score is chosen to be the evaluation metric for all our models. 
\begin{equation}
\label{eqn:F1Score}
\text{\emph{F1-score}} = \frac{2 \cdot {Precision}\cdot {Recall}}{{Precision}+{Recall}}
\end{equation}

Table \ref{tab:Task1aComparison} showcases the performance of the different models which performed well on the validation set. 
The RoBERTa model that was finally chosen after hyperparameter tuning achieved the highest score on the leaderboard among all teams participating in the subtask. The score obtained on the test set can be found in Table  \ref{tab:Task1aFinal}.

It can be seen in the results of the validation set and test for the ADE class that the recall is 0.92 for the validation set and 0.752 for the test set. The results show that the model has learnt features for classifying ADE samples from a small amount of data. Although it might classify some amount of NoADE tweets incorrectly as evidenced by the low precision, the greater number of correctly classified ADE tweets aligns with our objective of classifying the maximum number of ADE tweets correctly as possible so that we don't lose valuable information about adverse drug effects that might be found. Our model achieved a significantly higher recall than the median of all other teams (Table \ref{tab:Task1aFinal}), indicating that a majority of ADE tweets are correctly classified. \par


\subsection{Task-1b: ADE Span Detection}
The goal of this subtask is to detect the text span of reported ADEs in tweets.

\subsubsection{Data and Pre-processing}
The given dataset consisted of 1,712 spans across 1,234 tweets. For the purpose of better training of the model, all tweets with duplicate or overlapping spans were manually removed. The decision to do this manually was to ensure that spans providing better context were kept instead of just individual words that would have been less helpful in discerning the meaning of the sentence.

\subsubsection{System Description}
The dataset was passed through a Named Entity Recognition (NER) pipeline made using the \verb|en_core_web_trf| model. The pipeline makes use of the \verb|roberta-base| model provided by Huggingface's Transformers library \cite{wolf2020huggingfaces}. The algorithm for extracting Adverse Effects from tweets is provided in Algorithm \ref{algo:overview}.

\begin{algorithm}
\SetAlgoLined
    \textbf{Input}: Input raw tweet $T$; \\
    \textbf{Output}: $Label$, Start char, End char, Span;\\
    Given ($T$), Classify the tweet with fairseq RoBERTa into ADE or NoADE; \\
    \If{$Label$ is ADE}
    {
        Perform NER on $T$ using spaCy NER pipeline; \\
        Return Start char, End char, Span;
    }
    \caption{Algorithm for Extraction of Adverse Drug Effects from Tweets}
    \label{algo:overview}
\end{algorithm}

\subsubsection{Experiments}
Two Named Entity Recognition (NER) pipelines, \verb|en_core_web_trf| (\url{https://spacy.io/models/en#en_core_web_trf}) and \verb|en_core_web_sm| (\url{https://spacy.io/models/en#en_core_web_sm})  were tried. The first is a RoBERTa based model while the second is a fast statistical entity recognition system trained on written web text that includes vocabulary, vectors, syntax and entities. 
After hyperparameter tuning, the transformer model was chosen. The model was trained for 150 epochs  with a dropout of 0.3, Adam optimizer \cite{kingma2014method} and a learning rate of 0.001 with (\textbeta$_1$, \textbeta$_2$)=(0.9, 0.999).

\begin{table*}[!htbp]
\centering
\begin{tabular}{|c|c|c|c|c|c|c|c|}
\hline
    \textbf{Model} & \textbf{Relaxed P} & \textbf{Relaxed R} & \textbf{Relaxed F1} & \textbf{Strict P} & \textbf{Strict R} & \textbf{Strict F1}\\
\hline
    \verb|en_core_web_sm| & 0.516 & \textbf{0.551} & 0.533 & 0.226 & 0.241 & 0.233 \\
\hline
     \verb|en_core_web_trf| & \textbf{0.561} & 0.529 & \textbf{0.544} & \textbf{0.275} & \textbf{0.253} & \textbf{0.263} \\
\hline
\end{tabular}
\caption{Scores on the Validation Set for the model for task 1b.}
\label{tab:Task1bValid}
\end{table*}

\begin{figure*}[h!]
    \centering
        \includegraphics[width=\textwidth]{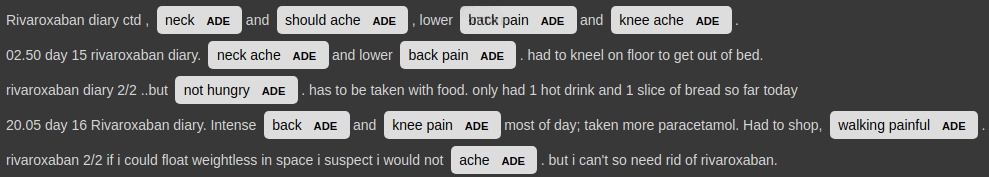}
        \includegraphics[width=\textwidth]{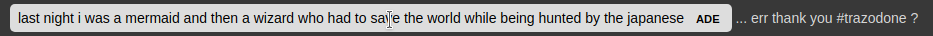}
    \caption{Example span extraction from TensorFlu's model for task 1b}
    \label{fig:examples}
\end{figure*}

\begin{table}[!htbp]
\centering
\resizebox{\columnwidth}{!}{\begin{tabular}{ |c|c|c|c|c|c|c| }
 \hline
  & \textbf{Precision} & \textbf{Recall} & \textbf{F1} \\ 
 \hline
 \texttt{\textbf{en\_core\_web\_trf}} & 0.493 & \textbf{0.505} & \textbf{0.50} \\
    \hline
    \texttt{en\_core\_web\_sm} & \textbf{0.521} & 0.458 & 0.49 \\
    \hline
 Median & 0.493 & 0.458 & 0.42 \\
    \hline
\end{tabular} }
\caption{Comparing our best-performing model to the median for task 1b.}
\label{tab:Task1bFinal}
\end{table}

\subsubsection{Results}
The models have been evaluated with two metrics, the Relaxed F1 score, and the Strict F1 score. The Relaxed metrics evaluate the scores for spans that have a partial or full overlap with the labels. The Strict metrics only evaluate the cases where the spans produced by the model perfectly match the span in the label.\par 
Table \ref{tab:Task1bValid} showcases the performance of both NER pipelines on the validation set. It can be observed that the RoBERTa model provides a higher F1 score than the statistical model and is able to make much more accurate classifications of the ADE class. The statistical model however provides a higher recall which indicates it has fewer false negatives and is thus misclassifying the ADE samples as NoADE less often. The RoBERTa model is however far superior to the statistical model when considering the strict F1 scores. This implies that it is able to produce a perfect span more often and has learnt a better representation of the data. \par The final test set result achieved by the model placed on the leaderboard was achieved by the RoBERTa based NER model. The results obtained by both models are compared to the median in Table \ref{tab:Task1bFinal}. The transformer pipeline provides a higher recall than the statistical pipeline thus showcasing the fact that a higher number of tweets were correctly classified as ADE while having overlapping spans. A few example images showing the performance of the entire adverse effect extraction pipeline  are provided in Figure \ref{fig:examples}.


\section{Task-6: Classification of COVID-19 tweets containing symptoms}
\label{Sec:task6}
The goal of this task is to classify tweets into 3 categories: (1) Self-reports (2) Non-personal reports (3) Literature/News mentions.

\subsection{Data and Pre-processing}
The SMM4H organizers released a training dataset consisting of 9,567 tweets and test data consisting of 6,500 tweets.
The training dataset consisted of 4,523 tweets with Literature/News mentions, 3,622 tweets with non-personal reports and 1,421 tweets with self-reports. There is very little class imbalance in the given dataset. Tokenization of tweets was done using VinAI's \verb|bertweet-base| tokenizer from the \textit{Huggingface} API \cite{wolf2020huggingfaces}. In order to use the BERTweet model, the tweets were normalized by converting user mentions into the @USER special token and URLs into the HTTPURL special token. The \textit{emoji} package was used to convert the emoticons into text. \cite{nguyen-etal-2020-bertweet}

\subsection{System Description}
BERTweet \cite{nguyen-etal-2020-bertweet} uses the same architecture as BERT base and the pre-training procedure is based on RoBERTa, \cite{liu2019roberta} for more robust performance, as it optimizes the BERT pre-training approach. BERTweet is optimized using Adam optimizer \cite{kingma2014method}, with a batch size of 7K and a peak learning rate of 0.0004, and is pre-trained for 40 epochs (using first 2 epochs for warming up the learning rate). The \verb|bertweet-covid19-base-uncased| model was used for our application, which has 135M parameters, and is trained on 845M English tweets and 23M COVID-19 English tweets.\par
For training the BERTweet model on our train dataset, (\url{https://github.com/VinAIResearch/BERTweet}) was used with number of labels set to 3.  
\begin{table*}[!ht]
\centering
\begin{tabular}{ |c|c|c|c|c|c|c| } 
 \hline
 \textbf{S.No.} & \textbf{Model} & \textbf{Arch} & \textbf{Label} & \textbf{Precision} & \textbf{Recall} & \textbf{F1} \\ 
 \hline
 \multirow{3}{*}{1.} & \multirow{3}{*}{RoBERTa} & \multirow{3}{*}{$BERT_{LARGE}$} & Lit-News mentions & 0.98 & 0.97 & 0.98 \\
    &  &    &   Nonpersonal reports   &   0.95    &   0.97    &   0.96 \\
    &  &    &   Self reports   &   0.97    &   0.96    &   0.97 \\
    \hline
\multirow{3}{*}{2.}  & \multirow{3}{*}{\textbf{BERTweet}}  & \multirow{3}{*}{$BERT_{BASE}$} & Lit-News mentions  & \textbf{0.99}  & 0.99  &   \textbf{0.99} \\
    &  &    &   Nonpersonal reports   &   \textbf{0.99}    &   \textbf{0.98}    &  \textbf{0.98} \\
    &  &    &   Self reports   &   0.97    &   \textbf{1.00}    &   \textbf{0.99} \\
    \hline
\multirow{3}{*}{3.}  & \multirow{3}{*}{DeBERTa}  & \multirow{3}{*}{$BERT_{BASE}$} & Lit-News mentions  & 0.95  & \textbf{1.00}   &   0.98 \\
    &  &    &   Nonpersonal reports   &   \textbf{0.99}    &   0.95    &   0.97   \\
    &  &    &   Self reports   &   \textbf{1.00}    &   0.95    &   0.97  \\
    \hline
\multirow{3}{*}{4.}  & \multirow{3}{*}{Covid-Twitter BERT}  & \multirow{3}{*}{$BERT_{LARGE}$} & Lit-News mentions  &  0.98 & 0.98  &    0.98\\
    &  &    &   Nonpersonal reports   &    0.97   &  0.97     &  0.97  \\
    &  &    &   Self reports   &    0.97   &  0.99     & 0.98   \\
    \hline
\multirow{3}{*}{5.} & \multirow{3}{*}{Majority Voting} & \multirow{3}{*}{NA} & Lit-News mentions & 0.98 & 0.99 & \textbf{0.99} \\
    &  &    &   Nonpersonal reports   &    0.98   &  0.97     &  0.97  \\
    &  &    &   Self reports   &    0.99   &  0.99     & \textbf{0.99}   \\
    \hline
\end{tabular}
\caption{Comparing different models used for task 6 on the \textbf{Validation Set}}
\label{tab:Task6Comparison}
\end{table*}

\subsection{Experiments}
A number of experiments were carried out to reach the optimal results for the task. Other models besides BERTweet were trained for the task such as RoBERTa \cite{liu2019roberta}, DeBERTa \cite{he2021deberta}, and  Covid-Twitter-BERT \cite{muller2020covidtwitterbert}. A majority voting ensemble with all 4 models was also evaluated. After a lot of tuning, BERTweet was found to be the best performing model on the dataset. \par 
The ideal hyperparameters for the model were found empirically following many experiments with the validation set. The best results were obtained with the following hyperparameters: the model was finetuned for 12 epochs with a batch size of 16; the learning rate was warmed up for 500 steps with a weight decay of 0.01.\par
Due to little class imbalance in the given dataset and pretrained BERT based models performing very well on classification tasks, almost all models achieved a relatively high F1-score.

\subsection{Results}
The results on the validation set for all the trained models are reported in Table  \ref{tab:Task6Comparison}. As mentioned in section \ref{sec:Equations} the models have been compared on the basis of Precision, Recall and F1-score. The best performing model as seen in Table \ref{tab:Task6Comparison} is BERTweet. The same model was also able to achieve an F1 score above the median on the test set as seen in Table \ref{tab:Task6Final}.

\begin{table}[!htbp]
\centering
\resizebox{\columnwidth}{!}{\begin{tabular}{ |c|c|c|c|c|c|c| } 
 \hline
  & \textbf{Precision} & \textbf{Recall} & \textbf{F1} \\ 
 \hline
 \textbf{BERTweet} & \textbf{0.9411} & \textbf{0.9411} & \textbf{0.94} \\
    \hline
 Median & 0.93235 & 0.93235 & 0.93 \\
    \hline
\end{tabular}}
\caption{Comparing our best-performing model to the median for task 6}
\label{tab:Task6Final}
\end{table}

\section{Conclusion}
In this work we have explored an application of RoBERTa to the task of classification, extraction and normalization of Adverse Drug Effect (ADE) mentions in English tweets and the application of BERTweet to the task of classification of tweets containing COVID-19 symptoms. We have based our selection of these models on a number of experiments we conducted to evaluate different models. Our experiments have shown that RoBERTa outperforms BioBERT, achieving state of the art results in ADE classification. For the second task, we found that BERTweet outperformed all the other models including an ensembling approach (majority voting). \par
We foresee multiple directions for future research. One possible improvement could be to use joint learning to deal with Task-1(a) and Task-1(b) simultaneously. 

\section{Acknowledgements}
For Task 1a, \textit{Sidharth R, Abhiraj Tiwari, Parthivi Choubey and Saisha Kashyap} contributed equally to the work. For Task 1b, \textit{Sidharth R, Abhiraj Tiwari, Sahil Khose, and Kumud Lakara} contributed equally to the work. All authors contributed equally to the work done for Task 6. We would also like to thank \textit{Mars Rover Manipal and Project Manas} for providing the necessary resources to train our models. 
\bibliographystyle{acl_natbib}
\bibliography{custom}

\end{document}